\documentclass[runningheads]{llncs}
\usepackage{graphicx}
\usepackage{amsmath,amssymb} 
\usepackage{color}
\usepackage{multirow}
\usepackage{array}
\usepackage[width=122mm,left=12mm,paperwidth=146mm,height=193mm,top=12mm,paperheight=217mm]{geometry}

\begin{document}
\pagestyle{headings}
\mainmatter
\def\ECCV16SubNumber{328}  

\title{DAVE: A Unified Framework for Fast Vehicle Detection and Annotation} 

\titlerunning{DAVE: A Unified Framework for Fast Vehicle Detection and Annotation}

\authorrunning{Yi Zhou, Li Liu, Ling Shao and Matt Mellor}

\author{Yi Zhou \inst{\dag}, Li Liu \inst{\dag}, Ling Shao \inst{\dag} \and Matt Mellor \inst{\ddag}}
\institute{{\dag} Northumbria University, Newcastle upon Tyne, NE1 8ST, UK \\ \inst{\ddag} Createc, Cockermouth, Cumbria, CA13 0HT, UK
\\ \{yi2.zhou, li2.liu, ling.shao\}@northumbria.ac.uk \ matt.mellor@createc.co.uk }

\maketitle

\begin{abstract}
Vehicle detection and annotation for streaming video data with complex scenes is an interesting but challenging task for urban traffic surveillance. In this paper, we present a fast framework of Detection and Annotation for Vehicles (DAVE), which effectively combines vehicle detection and attributes annotation. DAVE consists of two convolutional neural networks (CNNs): a fast vehicle proposal network (FVPN) for vehicle-like objects extraction and an attributes learning network (ALN) aiming to verify each proposal and infer each vehicle's pose, color and type simultaneously. These two nets are jointly optimized so that abundant latent knowledge learned from the ALN can be exploited to guide FVPN training. Once the system is trained, it can achieve efficient vehicle detection and annotation for real-world traffic surveillance data. We evaluate DAVE on a new self-collected UTS dataset and the public PASCAL VOC2007 car and LISA 2010 datasets, with consistent improvements over existing algorithms.
\end{abstract}

\section{Introduction and Related Work}
Automatic analysis of urban traffic activities is an urgent need due to essential traffic management and increased vehicle violations. Among many traffic surveillance techniques, computer vision-based methods have attracted a great deal of attention and made great contributions to realistic applications such as vehicle counting, target vehicle retrieval and behavior analysis. In these research areas, efficient and accurate vehicle detection and attributes annotation is the most important component of traffic surveillance systems.

Vehicle detection is a fundamental objective of traffic surveillance. Traditional vehicle detection methods can be categorized into frame-based and motion-based approaches~\cite{sivaraman2013looking,buch2011review}. For motion-based approaches, frames subtraction \cite{park2007video}, adaptive background modeling \cite{stauffer1999adaptive} and optical flow \cite{martinez2008driving,liu2013learning} are often utilized. However, some non-vehicle moving objects will be falsely detected with motion-based approaches since less visual information is exploited.  To achieve higher detection performance, recently, the deformable part-based model (DPM) \cite{felzenszwalb2010object} employs a star-structured architecture consisting of root and parts filters with associated deformation models for object detection. DPM can successfully handle deformable object detection even when the target is partially occluded. However, it leads to heavy computational costs due to the use of the sliding window precedure for appearance features extraction and classification.

With the wide success of deep networks on image classification \cite{krizhevsky2012imagenet,karpathy2014large,wu2016deep,wu2014leveraging,dong2015deep}, a Region-based CNN (RCNN) \cite{girshick2014rich} combines object proposals, CNN learned features and an SVM classifier for accurate object detection. To further increase the detection speed and accuracy, Fast RCNN \cite{girshick2015fast} adopts a region of interest (ROI) pooling layer and the multi-task loss to estimate object classes while predicting bounding-box positions. ``Objectness" proposal methods such as Selective Search \cite{uijlings2013selective} and Edgeboxes \cite{zitnick2014edge} can be introduced in RCNN and Fast RCNN to improve the efficiency compared to the traditional sliding window fashion. Furthermore, Faster RCNN \cite{ren2015faster} employs a Region Proposal Network (RPN) with shared convolutional features to enable cost-free effective proposals. All these deep models target general object detection. In our task, we aim for real-time detection of a special object type, vehicle.

Besides, in urban traffic surveillance, another interesting and valuable task is to extract more diverse information from detected vehicles - we call it vehicle attributes annotation. Each individual vehicle on the road has its special attributes: travel direction (i.e., pose), inherent color, type and other more fine-grained information with respect to the headlight, grille and wheel. It is extremely beneficial to annotate a target vehicle's attributes accurately. Lin et al. \cite{yang2014object} presents an auto-masking neural network for vehicle detection and pose estimation. In \cite{li2010vehicle}, an approach by vector matching of template is introduced for vehicle color recognition. In \cite{dong2014vehicle}, unsupervised convolutional neural network is adopted for vehicle type classification from frontal view images. However, independent analysis of different attributes makes visual information not well explored and the process inefficient, and little work has been done for annotating these vehicle attributes simultaneously. Actually, there exist strong correlations between these vehicle attributes learning tasks. For example, vehicle type classification based on visual structures is very dependent on the viewpoint. Therefore, we believe multi-task learning \cite{shao2015deeply,liu2016dap3d} can be helpful since such joint learning schemes can implicitly learn the common features shared by these correlated tasks. Moreover, a unified multi-attributes inference model can also significantly improve the efficiency.

\begin{figure}[t]
  \centering
  \includegraphics[width=1\textwidth]{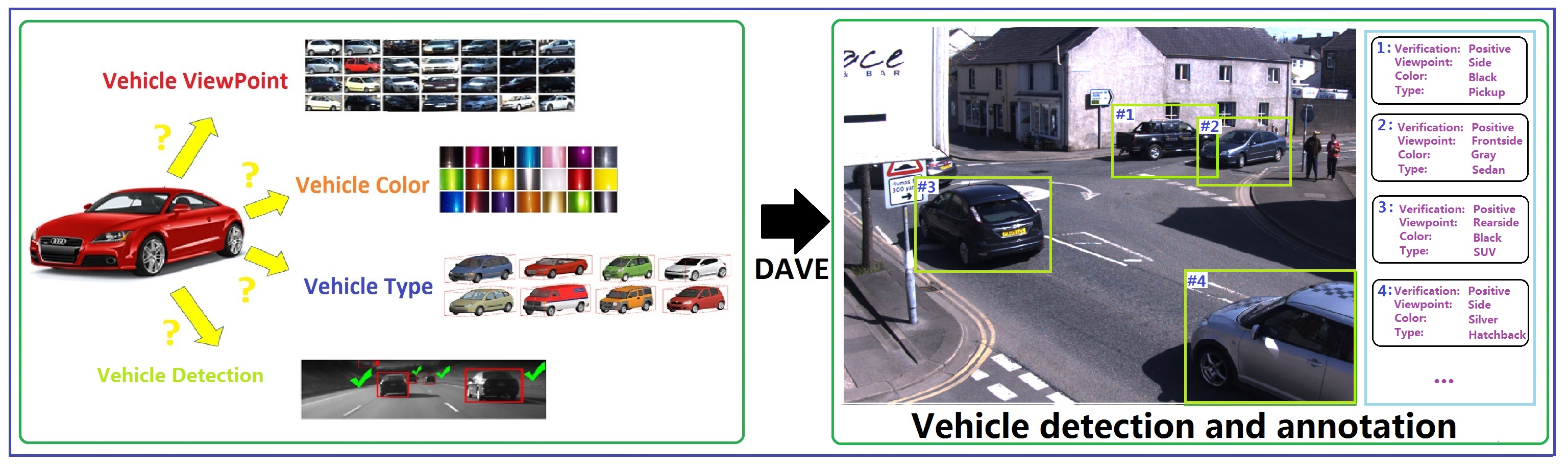}\\
  \caption{\scriptsize{
  \textbf{Illustration of DAVE}. Vehicle detection and corresponding pose, color and type annotation can be simultaneously achieved by DAVE as shown in the right sub-figure.}}
  \label{exampless}
  \vspace{-3ex}
\end{figure}

Inspired by the advantages and drawbacks of previous work,  in this paper, we propose a fast framework DAVE based on convolutional neural networks (CNNs), as shown in  Fig. 1, for vehicle detection and attributes annotation in urban traffic surveillance.   In particular, DAVE consists of two CNNs: fast vehicle proposal network (FVPN) and attributes learning network (ALN). The FVPN is a shallow\emph{ fully convolutional network} which is able to predict all the bounding-boxes of vehicle-like objects in real-time. The latter ALN configured with a very deep structure can precisely verify each proposal and infer pose, color and type information for positive vehicles, simultaneously.   It is noteworthy that informative features learned from the deep ALN can be regarded as latent data-driven knowledge to guide the training of the shallow FVPN, thus we bridge the ALN and FVPN with such knowledge guidance and jointly optimize these two CNNs at the same time in our architecture. In this way, more exhaustive vehicle descriptions learned from the ALN as helpful supervision benefits the FVPN with better performance. Once the joint training is finished, a two-stage inference scheme will be adopted for final vehicle annotation. The main contributions of our work are highlighted as follows:

\textbf{1.} We unify multiple vehicle-related tasks into one deep vehicle annotation framework DAVE which can effectively and efficiently annotate each vehicle's bounding-box, pose, color and type simultaneously.

\textbf{2.} Two CNNs proposed in our method, i.e., the Fast Vehicle Proposal Network (FVPN) and the vehicle Attributes Learning Network (ALN), are optimized in a joint manner by bridging two CNNs with latent data-driven knowledge guidance. In this way, the deeper ALN can benefit the performance of the shallow FVPN.

\textbf{3.} We introduce a new Urban Traffic Surveillance (UTS) vehicle dataset consisting of six $1920\times1080$ (FHD) resolution videos with varying illumination conditions and viewpoints.

\section{Detection and Annotation for Vehicles (DAVE)}
We unify vehicle detection and annotation of pose, color and type into one framework: DAVE. As illustrated in Fig. 2, DAVE consists of two convolutional neural networks called fast vehicle proposal network (FVPN) and attributes learning network (ALN), respectively. FVPN aims to predict all the positions of vehicle-like objects in real-time.  Afterwards, these vehicle proposals are passed to the ALN to simultaneously verify all the positive vehicles and infer their corresponding poses, colors and types. In the training phase, FVPN and ALN are optimized jointly, while two-stage inference is used in the test phase. Specifically, training our DAVE is inspired by \cite{hinton2015distilling} that knowledge learned from solid deep networks can be distilled to teach shallower networks. We apply latent data-driven knowledge from the deep ALN to guide the training of the shallow FVPN.
This method proves to be able to enhance the performance of the FVPN to some extent through experiments. The architecture of FVPN and ALN will be described in the following subsections.

\begin{figure}[t]
  \centering
  \includegraphics[width=0.975\textwidth]{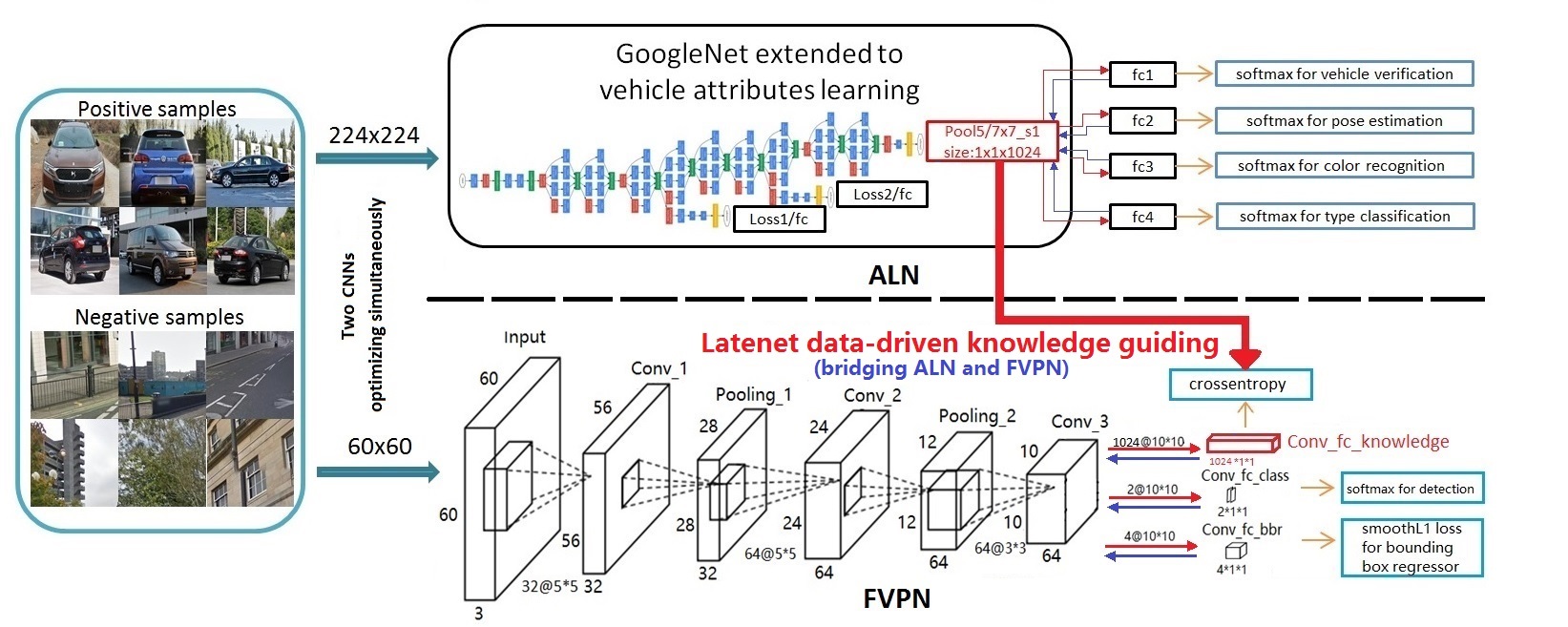}\\
  \caption{
  \textbf{Training Architecture of DAVE.} Two CNNs: FVPN and ALN are simultaneously optimized in a joint manner by bridging them with latent data-driven knowledge guidance.}
  \label{exampless}
\end{figure}

\subsection{Fast Vehicle Proposal Network (FVPN)}
Searching the whole image to locate potential vehicle positions in a sliding window fashion is prohibitive for real-time applications. Traditional object proposal methods are put forward to alleviate this problem, but thousands of proposals usually contain numerous false alarms and duplicate predictions which heavily lower the efficiency. Particularly for one specific object, we expect very fast and accurate proposal performance can be achieved.

Our proposed fast vehicle proposal network (FVPN) is a shallow \emph{fully convolutional network}, which aims to precisely localize all the vehicle-like objects in real-time. We are interested in exploring whether or not a small scale CNN is enough to handle the single object proposal task. A schematic diagram of the FVPN is depicted in the bottom part of Fig. 2. The first convolutional layer (\emph{conv\_1}) filters the $60\times60$ training images with 32 kernels of size $5\times5$. The second convolutional layer (\emph{conv\_2}) takes as input the feature maps obtained from the previous layer and filters them with 64 kernels of size $5\times5$. Max pooling and Rectified Linear Units (ReLU) layers are configured after the first two convolutional layers. The third convolutional layer (\emph{conv\_3}) with 64 kernels of size $3\times3$ is branched into three sibling  $1\times1$ convolutional layers transformed by traditional fully connected layers. In detail, \emph{Conv\_fc\_class} outputs softmax probabilities of positive samples and the background; \emph{Conv\_fc\_bbr} encodes bounding-box coordinates for each positive sample; \emph{Conv\_fc\_knowledge} is configured for learning latent data-driven knowledge distilled from the ALN, which makes the FVPN be trained with more meticulous vehicle features.  Inspired by \cite{long2015fully}, these $1\times1$ convolutional layers can successfully lead to differently purposed heatmaps in the inference phase. This property will directly achieve real-time vehicle localization from whole images/frames by our FVPN.

We employ different loss supervision layers for three corresponding tasks in the FVPN. First, discrimination between a vehicle and the background is a simple binary classification problem. A softmax loss layer is applied to predict vehicle confidence, $p^c=\{p_{ve}^c,p_{bg}^c\}$. Besides, each bounding-box is encoded by 4 predictions: $x$, $y$, $w$ and $h$. $x$ and $y$ denote the left-top coordinates of the vehicle position, while $w$ and $h$ represent the width and height of the vehicle size. We normalize all the 4 values relative to the image width and height so that they can be bounded between 0 and 1. Note that all bounding boxes' coordinates are set as \emph{zero} for background samples. Following \cite{girshick2015fast}, a smooth L1 loss layer is used for bounding-box regression to output the refined coordinates vector, $loc=(\hat{x},\hat{y},\hat{w},\hat{h})$. Finally, for guiding with latent data-driven knowledge of an N-dimensional vector distilled from a deeper net, the cross-entropy loss is employed for $p^{know}=\{p_0^{know}\ldots p_{N-1}^{know}\}$.

We adopt a multi-task loss $L_{FVPN}$ on each training batch to jointly optimize binary classification of a vehicle against background, bounding-box regression and learning latent knowledge from a deeper net as the following function:

\begin{equation}
\label{eq:e1}
\small
L_{FVPN}(loc, p^{bic},p^{know})=L_{bic}(p^{bic})+\alpha L_{bbox}(loc)+\beta L_{know}(p^{know}),
\end{equation}

where $L_{bic}$ denotes the softmax loss for binary classification of vehicle and background. $L_{bbox}$ indicates a smooth $\ell_1$ loss defined in \cite{girshick2015fast} as:

\begin{equation}
\label{eq:e1}
\small
L_{bbox}(loc)=f_{L1}(loc-loc_{t}),\ \mathrm{where}\   f_{L1}(x) = \left\{\begin{array}{l}
                       0.5x^{2},~  \text{if}\ |x|<1 \\
                       |x|-0.5,~  \text{otherwise}
                     \end{array}
  \right.
\end{equation}

Furthermore, the cross entropy loss $L_{know}$ is to guide the training of the FVPN by a latent N-dimensional feature vector $t^{know}$ learned from a more solid net, which is defined as:
\begin{equation}
\small
L_{know}(p^{know})=-\frac{1}{N}\sum_{i}^{N}t^{know}_{i}\log p^{know}_{i}+(1-t^{know}_{i})\log (1-p^{know}_{i}).
\end{equation}

It is noteworthy that  a bounding-box for the background is meaningless in the FVPN back-propagation phase and will cause training to diverge early \cite{redmon2015you}, \emph{thus we set $\alpha=0$ for background samples, otherwise $\alpha=0.5$.} Besides, $\beta$ is fixed to $0.5$.

\subsection{Attributes Learning Network (ALN)}

Modeling vehicle pose estimation, color recognition and type classification separately is less accurate and inefficient. Actually, relationships between these tasks can be explored, so that designing a multi-task network is beneficial for learning shared features which can lead to extra performance gains. The attribute learning network (ALN) is a unified network to verify vehicle candidates and annotate their poses, colors and types. The network architecture of the ALN is mainly inspired by the GoogLeNet \cite{szegedy2015going} model. Specifically, we design the ALN by adding 4 fully connected layers to extend the GoogLeNet into a multi-attribute learning model. The reason to adopt such a very deep structure here is because vehicle annotation belongs to fine-grained categorization problems and a deeper net has more powerful capability to learn representative and discriminative features. Another advantage of the ALN is its high-efficiency inherited from the GoogLeNet which has lower computation and memory costs compared with other deep nets such as the VGGNet \cite{simonyan2014very}.

The ALN is a multi-task network optimized with four softmax loss layers for vehicle annotation tasks. Each training image has four labels in $V$, $P$, $C$ and $T$. $V$ determines whether a sample is a vehicle. If $V$ is a true vehicle, the remaining three attributes $P$, $C$ and $T$ represent its pose, color and type respectively. However, if $V$ is the background or a vehicle with a catch-all\footnote{``Catch-all" indicates other undefined types and colors which are not included in our training model.} type or color, $P$, $C$ and $T$ are set as \emph{zero} denoting attributes are unavailable in the training phase. The first softmax loss layer $L_{verify} (p^V)$ for binary classification (vehicle vs. background) is the same as $L_{bic} (p^c)$ in the FVPN. The softmax loss $L_{pose} (p^P)$, $L_{color} (p^C)$ and $L_{type} (p^T)$ are optimized for pose estimation,  color recognition and vehicle type classification respectively, where  $p^P=\{p_1^P,\ldots,p_{np}^P\}$, $p^C=\{p_1^C,\ldots,p_{nc}^C\}$ and $p^T=\{p_1^T,\ldots, p_{nt}^T\}$. \{$np$, $nc$, $nt$\} indicate the number of vehicle poses, colors and types respectively.    The whole loss function is defined as follows:

\begin{equation}
\label{eq:e1}
\small
L_{ALN}(p^{V},p^{P},p^{C},p^{T})=L_{verify}(p^{V})+\lambda_{1} L_{pose}(p^{P})+\lambda_{2} L_{color}(p^{C})+\lambda_{3} L_{type}(p^{T}),
\end{equation}

where all the four sub loss functions are softmax loss for vehicle verification (\emph{\textbf{``verification" in this paper means confirming whether a proposal is vehicle}}), pose estimation, color recognition and type classification. \emph{Following the similar case of $\alpha$ in Eq. (1), parameters $\{\lambda_1,\lambda_2,\lambda_3\}$ are all fixed as 1 for the positive samples, otherwise as 0 for the background.}

\subsection{Deep Nets Training}

\subsubsection{Training Dataset and Data Augmentation}
We adopt the large-scale CompCars dataset \cite{yang2015large} with more than 100,000 web-nature data as the positive training samples which are annotated with tight bounding-boxes and rich vehicle attributes such as pose, type, make and model. In detail, the web-nature part of the CompCars dataset provides five viewpoints as \emph{front}, \emph{rear}, \emph{side}, \emph{frontside} and \emph{rearside}, twelve vehicle types as \emph{MPV, SUV, sedan, hatchback, minibus, pickup, fastback, estate, hardtop-convertible, sports, crossover} and \emph{convertible}. To achieve an even training  distribution, we discard less common vehicle types with few training images and finally select six types with all the five viewpoints illustrated in Fig. 3(a) to train our model. Besides, since color is another important attribute of a vehicle, we additionally annotate colors on more than 10,000 images with five common vehicle colors as \emph{black, white, silver, red} and \emph{blue} to train our final model. Apart from positive samples, about 200,000 negative samples without any vehicles are cropped from Google Street View Images to compose our training data.

For data augmentation, we first triple the training data with increased and decreased image intensities for making our DAVE more robust under different lighting conditions. In addition, image downsampling up to 20\% of the original size and image blurring are introduced to enable that detected vehicles with small sizes can be even annotated as precisely as possible.

\subsubsection{Jointly Training with Latent Knowledge Guidance}
The entire training structure of DAVE is illustrated in Fig. 2. We optimize the FVPN and the ALN jointly but with different sized input training data at the same time. The input resolution of the ALN is $224\times224$ for fine-grained vehicle attributes learning, while it is decreased to $60\times60$ for the FVPN to fit smaller scales of the test image pyramid for efficiency in the inference phase. In fact, the resolution of $60\times60$ can well guarantee the coarse shape and texture of a vehicle is discriminative enough against the background. Besides, another significant difference between the ALN and the FVPN  is that input vehicle samples for the ALN are tightly cropped, however, for the FVPN, uncropped vehicles are used for bounding-box (labeled as $loc_t$ in Eq. (2)) regressor training.

\begin{figure*}[t]
  \centering
  \begin{tabular}{cc}
    \includegraphics[width=0.49\textwidth,height=0.22\textheight]{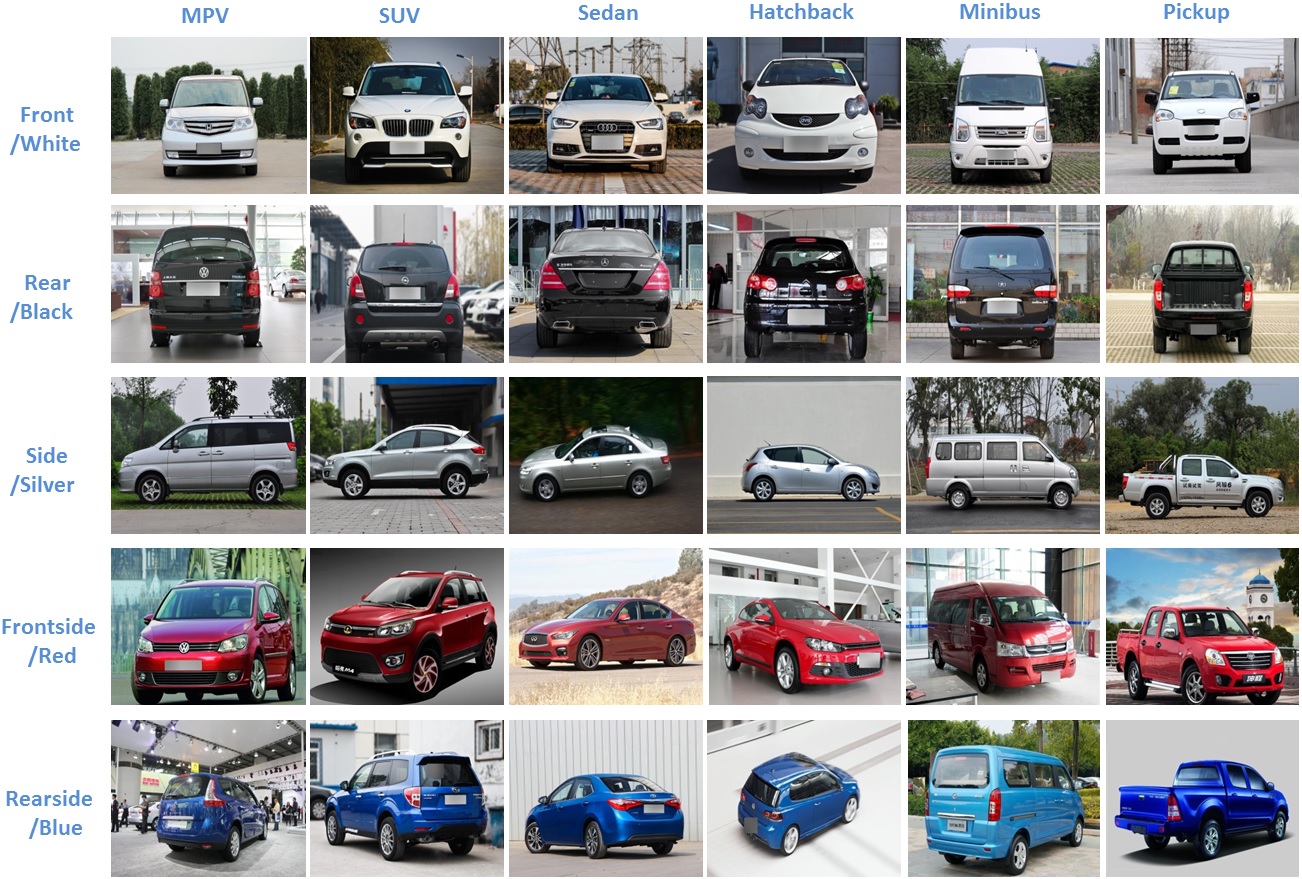} &\includegraphics[width=0.49\textwidth,height=0.22\textheight]{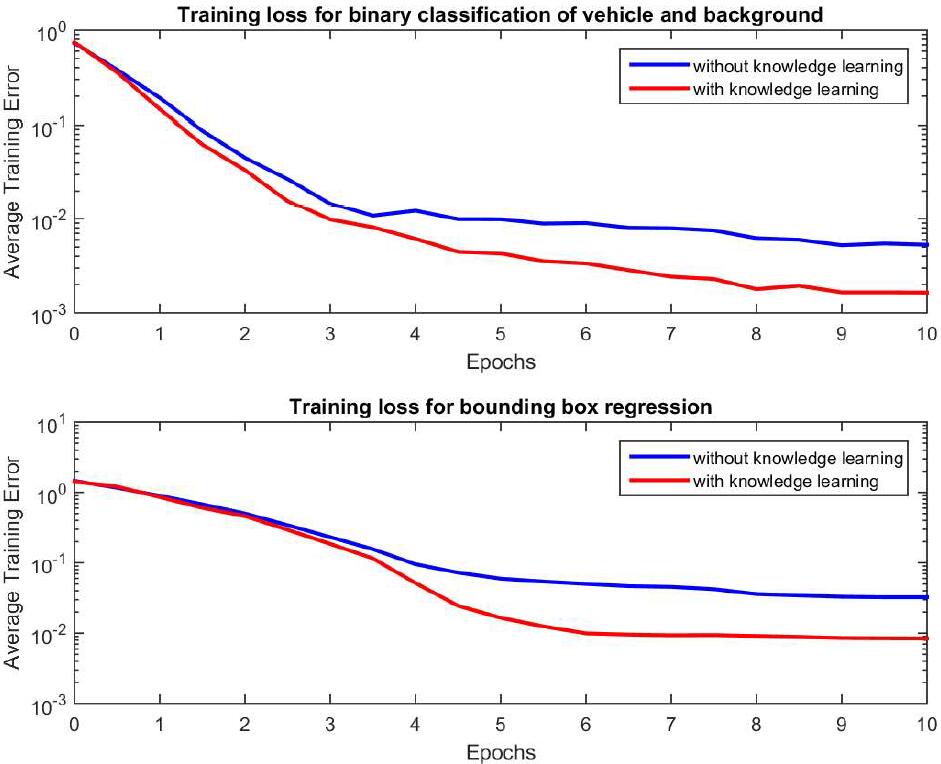}\\
    \footnotesize
    (a) &\footnotesize (b) \\
  \end{tabular}
  \caption{
  \textbf{}(a) Examples of training data (columns indicate vehicle types, while rows indicate poses and colors), (b) Training loss with/without knowledge learning.}
  \label{nasaacc}
\end{figure*}

The pre-trained GoogLeNet model for 1000-class ImageNet classification is used to initialize all the convolutional layers in the ALN, while the FVPN is trained from scratch.  A 1024-dimensional feature vector of the \emph{pool5/7$\times$7\_s1} layer in the ALN, which can exhaustively describe a vehicle, is extracted as the latent data-driven knowledge guidance to supervise the same dimensional \emph{Conv\_fc\_knowledge} layer in the FVPN by cross entropy loss.

We first jointly train ALN and FVPN for about 10 epochs on the selected web-nature data that only contains pose and type attributes from the CompCars. In the next 10 epochs, we fine-tune the models by a subset with our complemented color annotations. Throughout the training process, we set the batch size as 64, and the momentum and weight decay are configured as 0.9 and 0.0002, respectively. Learning rate is scheduled as $10^{-3}$ for the first 10 epochs and $10^{-4}$ for the second 10 epochs. To make our method more convincing, we train two models with and without knowledge guidance, respectively. During training, we definitely discover that knowledge guidance can indeed benefit training the shallow FVPN to obtain lower training losses. Training loss curves for the first 10 epochs are depicted in Fig. 3(b).

\begin{figure} [t]
  \centering
  \includegraphics[width=0.975\textwidth]{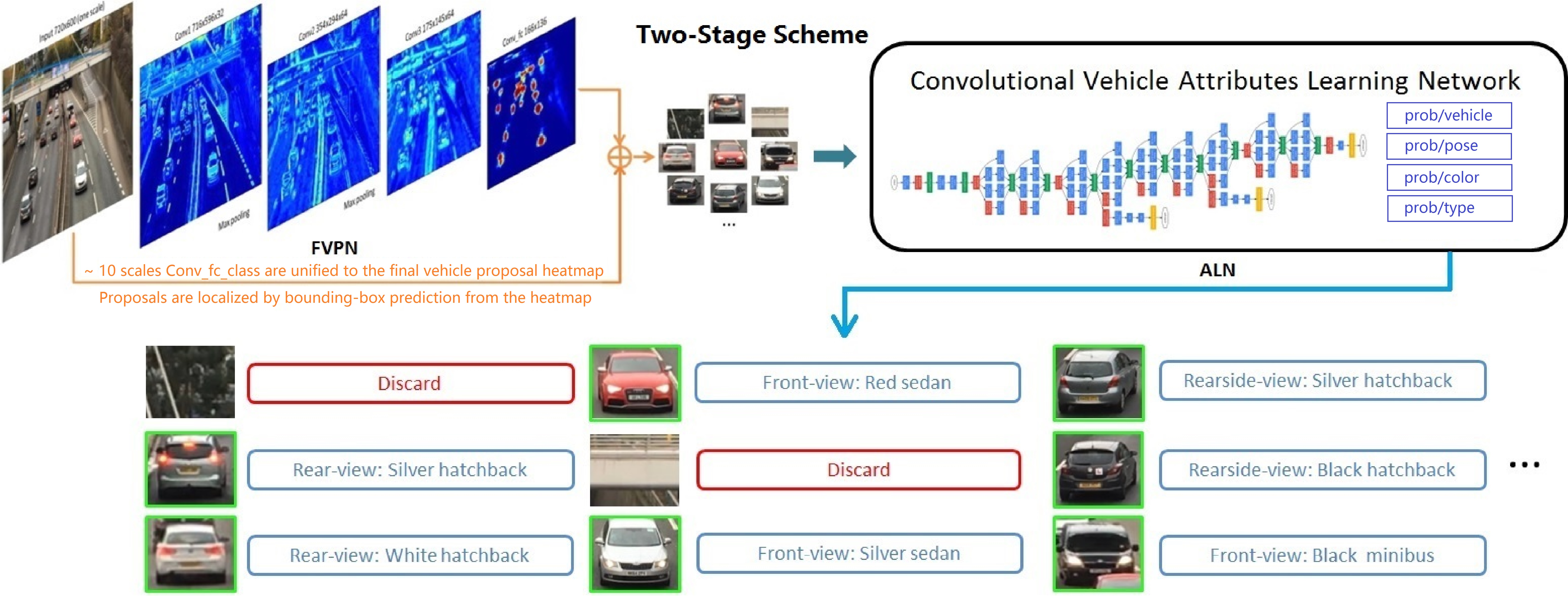}\\
  \caption{
  \textbf{A two-stage inference phase of DAVE.} Vehicle proposals are first obtained from FVPN in real-time. Afterwards, we use ALN to verify each proposal and annotate each positive one with the vehicle pose, color and type.}
  \label{exampless}
\end{figure}

\subsection{Two-stage Deep Nets Inference}
Once the joint training is finished, a two-stage scheme is implemented for inference of DAVE. First, the FVPN takes as input the 10-level test image Gaussian pyramid. For each level, the FVPN is operated over the input frame to infer \emph{Conv\_fc\_class} and \emph{Conv\_fc\_bbr} layers as corresponding heatmaps. All the 10 \emph{Conv\_fc\_class} heatmaps are unified into one map by rescaling all the channels to the largest size among them and keeping the maximum along channels, while the index of each maximum within 10 channels is used to obtain four unified \emph{Conv\_fc\_bbr} heatmaps (10 levels by similar rescaling). After unifying different levels \emph{Conv\_fc\_class} heatmaps into the final vehicle proposal score map, we first filter the score map with threshold $thres$ to discard low hot spots, and then local peaks on the map are detected by a circle scanner with tuneable radius $r$. In all our experiments, $r = 8$ and $thres = 0.5$ are fixed. Thus, these local maximal positions are considered as the central coordinates of proposals, ($\hat{x_i}$ ,$\hat{y_i}$). Coarse width and height of each proposal can be simply predicted based on the bounding-box of its corresponding hot spot centered on each local peak. If one hot spot contains multiple peaks, the width and height will be shared by these peaks (i.e. proposals). For preserving the complete vehicle body, coarse width and height are multiplied by fixed parameter $m = 1.5$ to generate ($\hat{w}_i^{nobbr}$ ,$\hat{h}_i^{nobbr}$). Thus, a preliminary bounding-box can be represented as ($\hat{x_i}$,$\hat{y_i}$,$\hat{w}_i^{nobbr}$ ,$\hat{h}_i^{nobbr}$). Finally, bounding-box regression offset values (within [0,1]) are extracted from four unified heatmaps of \emph{Conv\_fc\_bbr} at those coordinates ($\hat{x_i}$,$\hat{y_i}$) to obtain the refined bounding-box.

Vehicle proposals inferred from the FVPN are taken as inputs into the ALN. Although verifying each proposal and annotation of attributes are at the same stage, we assume that verification has a higher priority. For instance, in the inference phase, if a proposal is predicted as a positive vehicle, it will then be annotated with a bounding-box and inferred pose, color and type. However, a proposal predicted as the background will be neglected in spite of its inferred attributes. Finally, we perform non-maximum suppression as in RCNN \cite{girshick2014rich} to eliminate duplicate detections. The full inference scheme is demonstrated in Fig. 4. At present, it is difficult to train a model that has the capability to annotate all the vehicles with enormously rich vehicle colors and types. During inference, a vehicle with untrained colors and types is always categorized into similar classes or a catch-all ``others" class, which is a limitation of DAVE. In future work, we may expand our training data to include more abundant vehicle classes.

\section{Experiments and Results}
In this section, we evaluate our DAVE for detection and annotation of pose, color and type for each detected vehicle. Experiments are mainly divided as two parts: vehicle detection and attributes learning. DAVE is implemented based on the deep learning framework Caffe \cite{jia2014caffe} and run on a workstation configured with an NVIDIA TITAN X GPU.

\subsection{Evaluation of Vehicle Detection}
To evaluate vehicle detection, we train our models using the large-scale CompCars dataset as mentioned before, and test on three other vehicle datasets. We collect a full high definition ($1920\times1080$) Urban Traffic Surveillance (UTS) vehicle dataset with six videos which were captured from different viewpoints and illumination conditions. Each video sequence contains 600 annotated frames. To be more convincing, we also compare our method on two other public datasets: the PASCAL VOC2007 car dataset \cite{everingham2010pascal} and the LISA 2010 dataset \cite{sivaraman2010general} with four competitive models: DPM \cite{felzenszwalb2010object}, RCNN \cite{ren2015faster}, Fast RCNN \cite{girshick2015fast} and Faster RCNN \cite{ren2015faster}. These four methods obtain state-of-the-art performances on general object detection and the codes are publicly available. We adopt the trained car model in voc-release5 \cite{voc-release5} for DPM, while train models (VGG-16 version) for RCNN, Fast RCNN and Faster RCNN ourselves on the CompCars dataset to implement vehicle detection as our DAVE. The vehicle detection evaluation criterion is the same as PASCAL object detection \cite{everingham2010pascal}. Intersection over Union (IoU) is set as 0.7 to assess correct localization.

\subsubsection{Testing on the UTS dataset}
Since our FVPN can be considered as a high-accuracy proposal model for a single specific object, we test it independently. Then, results from vehicle verification by the deeper ALN (i.e., FVPN+verify in Fig. 5) are shown as our final accuracies. The detection accuracy as average precision (AP) and speed as frames-per-second (FPS) are compared in the left column of Table 1. Our model outperforms all the other methods with obvious improvements. Specifically, our method obtains an increased AP of 3.03\% compared to the best model Faster RCNN, and the shallow proposal network FVPN independently achieves 57.28\% which is only lower than Faster RCNN. The other two deep models, RCNN and Fast RCNN, do not produce satisfactory results mainly due to the low-precision proposals extracted by Selective Search \cite{uijlings2013selective}. Mixture-DPM with bounding-box prediction (MDPM-w-BB \cite{felzenszwalb2010object}) significantly improve the performance compared to MDPM-w/o-BB \cite{felzenszwalb2010object} by 10.77\%. For the evaluation of efficiency, our FVPN with a shallow and thin architecture can achieve real-time performance with 30 fps on FHD video frames. Although the deeper ALN slows the entire detection framework, 2 fps performance still shows competitive speed with Faster RCNN (4 fps).

We also test the FVPN trained without knowledge guidance, the AP decreases by 2.37\%, which explains the significant advancement of knowledge guidance. In addition, experiments are carried out to demonstrate that bounding-box regression can be helpful with the AP increased by 0.96\%.

\begin{table}
\center
\newcommand{\tabincell}[2]{\begin{tabular}{@{}#1@{}}#2\end{tabular}}
\caption{\small
\textbf{Vehicle detection AP (\%) and speed (fps) comparison on the UTS, PASCAL VOC2007 and LISA 2010 datasets}}
\label{table:t1}
\resizebox{1\textwidth}{!}{
\begin{tabular}{|c||c|c||c|c||c|c|}
\cline{1-7}
\multirow{3}{*}{\textbf{\tabincell{c}{Methods}}}&\multicolumn{2}{c||}{\textbf{UTS}} &\multicolumn{2}{c||}{\textbf{PASCAL VOC2007}}&\multicolumn{2}{c|}{\textbf{LISA 2010}} \\
\cline{2-7}
& \textbf{\tabincell{c}{Average\\ Precision (AP)}} & \textbf{\tabincell{c}{Processing\\ Speed (fps)}} & \textbf{\tabincell{c}{Average\\ Precision (AP)}} &\textbf{\tabincell{c}{Processing\\ Speed (fps)}}& \textbf{\tabincell{c}{Average\\ Precision (AP)}} &\textbf{\tabincell{c}{Processing\\ Speed (fps)}}\\
\hline
\hline
MDPM-w/o-BB&41.96\% &0.25 &48.44\% &1.25 &63.61\% &0.7 \\
\hline
MDPM-w-BB&52.73\% &0.2 &57.14\% &1.25 &72.89\% &0.7 \\
\hline
RCNN&44.87\% &0.03 &38.52\% &0.08 &55.37\% &0.06 \\
\hline
FastRCNN&51.58\% &0.4 &52.95\% &0.5 &53.37\% &0.5 \\
\hline
FasterRCNN&59.82\% &\textbf{4} &63.47\% &\textbf{6} &77.09\% &\textbf{6} \\
\hline
\hline
FVPN-w/o-knowledge guide &54.91\% &30 &58.12\% &46 &72.37\% &42 \\
\hline
FVPN-w/o-bbr&56.32\% &30 &58.93\% &46 &71.82\% &42 \\
\hline
\textbf{FVPN}&57.28\% &\textbf{30} &60.05\% &\textbf{46} &73.46\% &\textbf{42} \\
\hline
\textbf{FVPN+Verify}&\textbf{62.85\%} &2 &\textbf{64.44\% }&4 &\textbf{79.41\%} &4 \\
\hline
\end{tabular}
}\tiny
\\``bbr" indicates the bounding-box regressor used in our model, while ``BB" denotes bounding-box prediction used in DPM model. ``w" and ``w/o" are the abbreviations of ``with" and ``without", respectively. ``Verify" denotes the vehicle verification in the ALN.
\vspace{-4ex}
\end{table}

\subsubsection{Testing on the PASCAL VOC2007 car dataset and the LISA 2010 dataset}
To make our methods more convincing, we also evaluate on two public datasets. All the images containing vehicles in the trainval and test sets (totally 1434 images) in the PASCAL VOC 2007 dataset are extracted to be evaluated. In addition, the LISA 2010 dataset contains three video sequences with low image quality captured from a on-board camera.
All the results are shown in the middle and right columns of Table 1. For the PASCAL VOC2007 dataset, our model (FVPN+verify) achieves 64.44\% in AP, which outperforms MDPM-w-BB, RCNN, FastRCNN and Faster RCNN by 7.3\%, 25.92\%, 11.49\% and 0.97\%, respectively. Likewise, our FVPN can even obtain a high AP of 60.05\%. For the LISA 2010 dataset, the highest accuracy of 79.41\% by our model beats all other methods as well. Therefore, it demonstrates that our method is able to stably detect vehicles with different viewpoints, occlusions and image qualities.

Fig. 5 presents the precision-recall curves of all the compared methods on UTS, PASCAL VOC2007 car dataset and the LISA 2010 dataset, respectively. From all these figures, we can further discover that, for all three datasets, FVPN+Verify achieves better performance than other detection methods by comparing Area Under the Curve (AUC). Besides, some qualitative detection results including successful and failure cases are shown in Fig. 6. It can be observed that the FVPN cannot handle highly occluded cases at very small sizes, since local peaks on the FVPN heatmap for vehicles in those cases will be overlapped. The similar situation also exists in most of the deep networks based detection approaches \cite{girshick2014rich,girshick2015fast,redmon2015you,ren2015faster}.

\begin{figure*} [t]
  \centering
  \begin{tabular}{ccc}
    \includegraphics[width=0.33\textwidth,height=0.18\textheight]{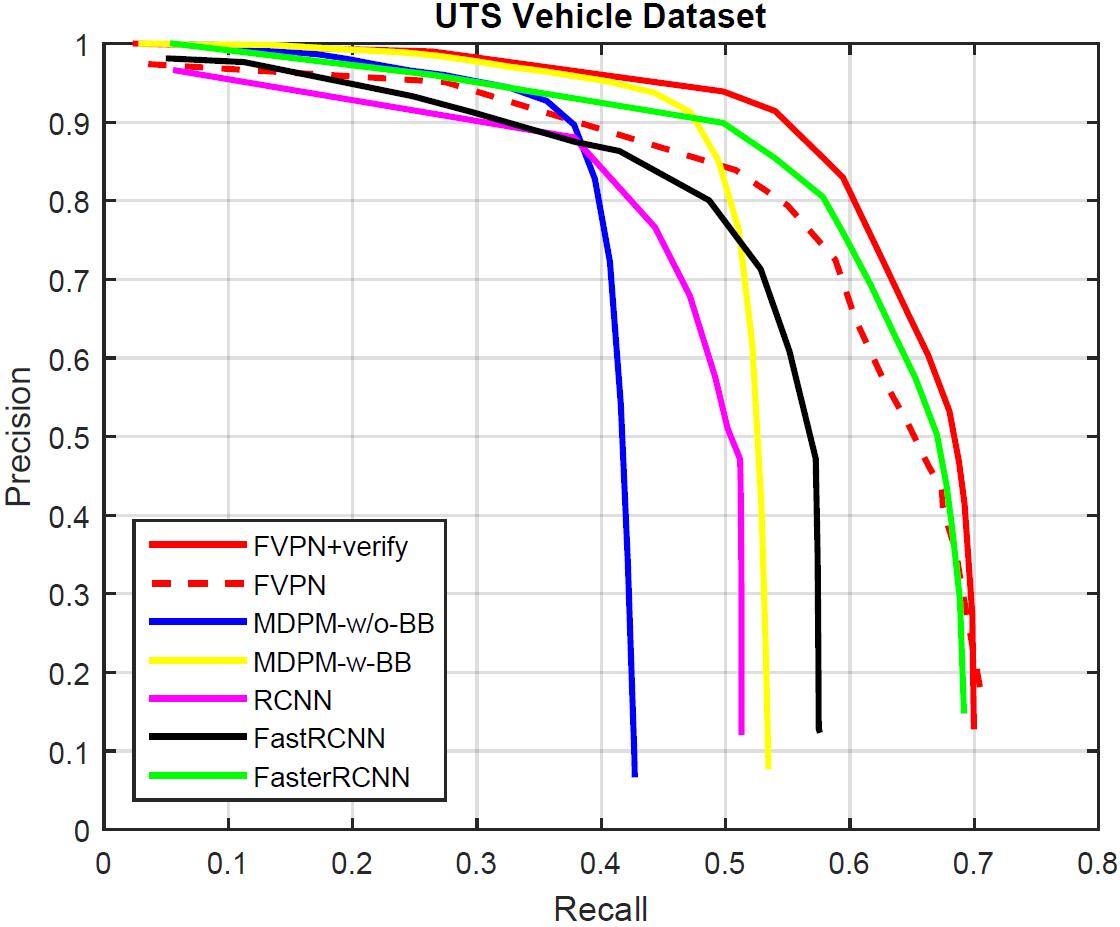} &\includegraphics[width=0.33\textwidth,height=0.18\textheight]{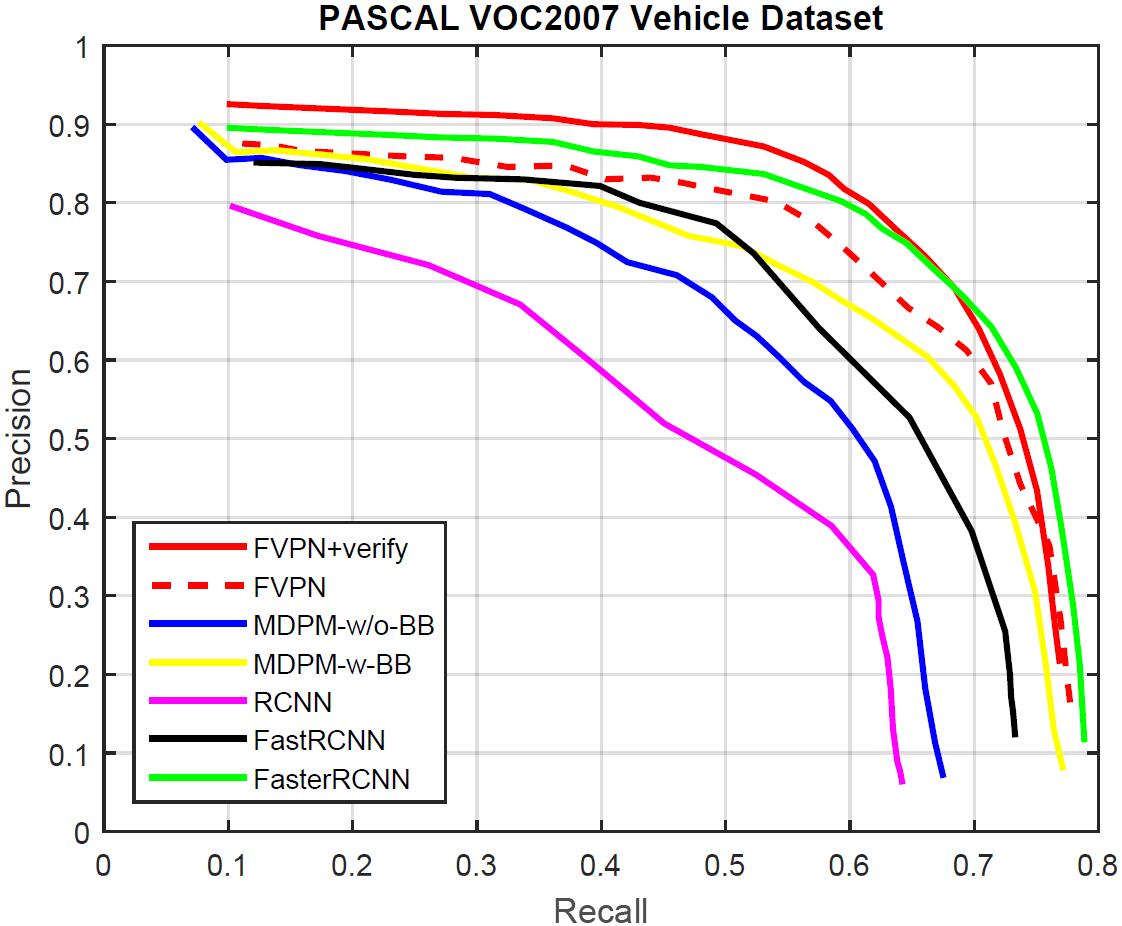}&\includegraphics[width=0.33\textwidth,height=0.18\textheight]{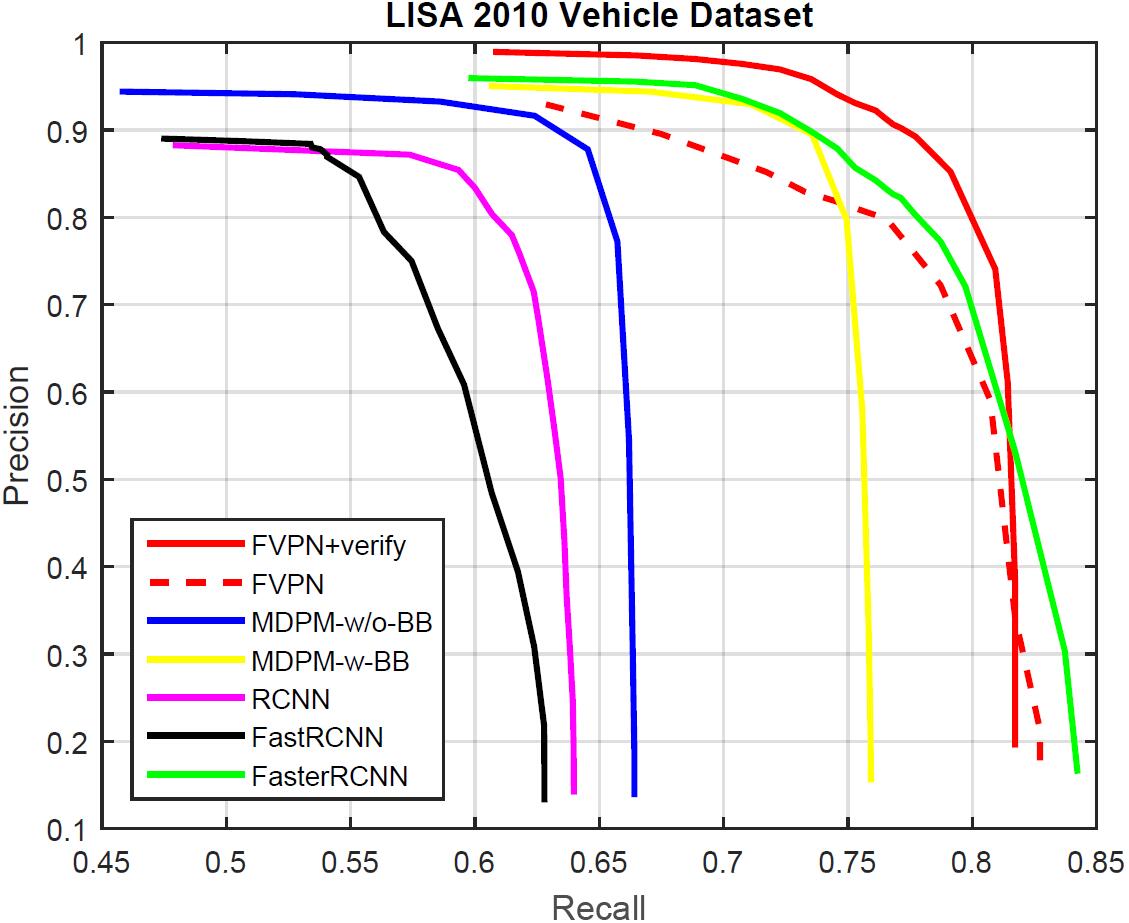}\\
    \footnotesize
    (a) &\footnotesize (b)  &\footnotesize (c)\\
  \end{tabular}
  \caption{
  \textbf{}Precision-recall curves on three vehicle datasets}
  \label{nasaacc}
\end{figure*}

\begin{figure} [t]
  \centering
  \includegraphics[width=0.95\textwidth]{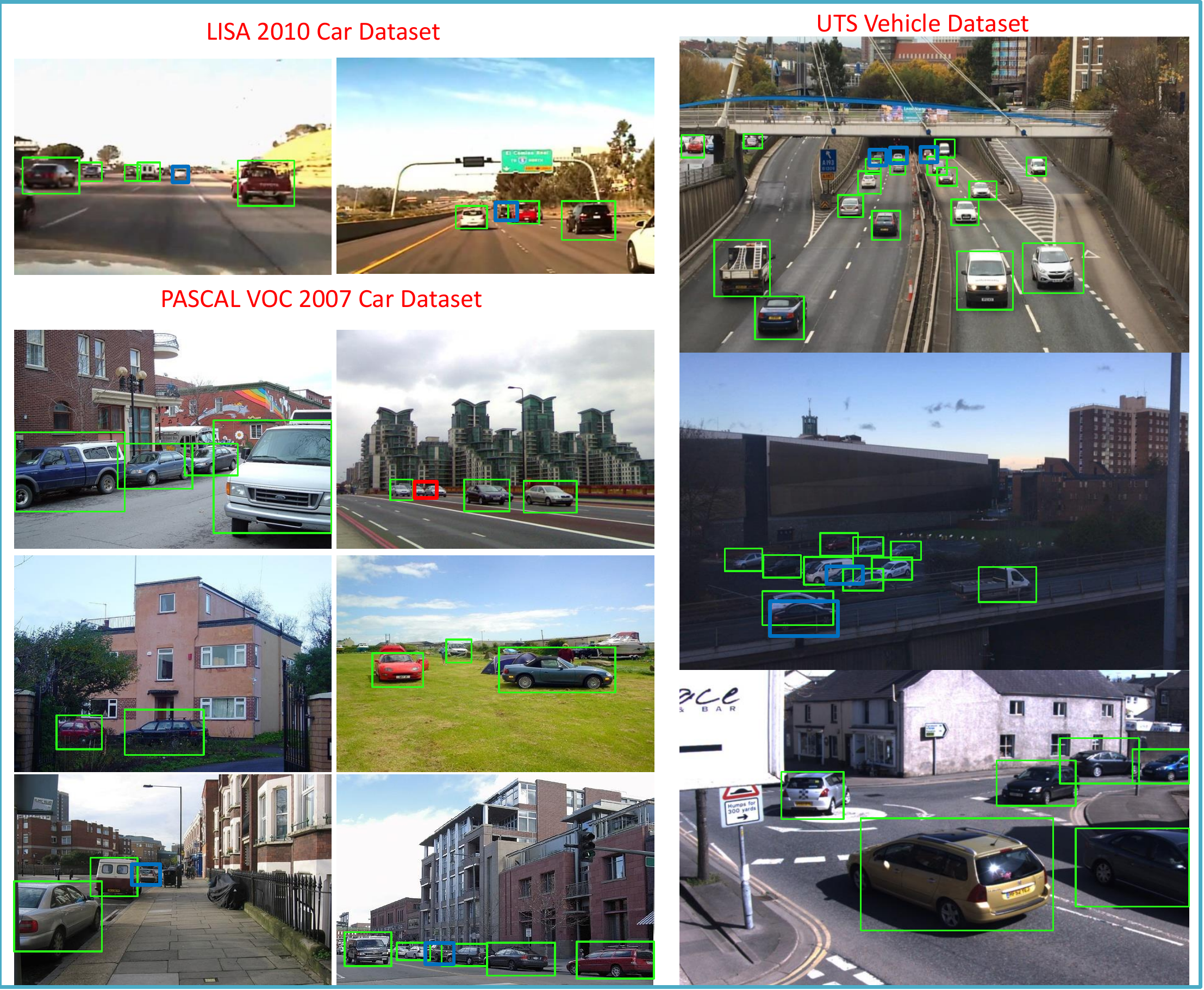}\\
  \caption{
  \textbf{}Examples of successful and failure cases for detection (a green box denotes correct localization, a red box denotes false alarm and a blue box denotes missing detection)}
  \label{exampless}
\end{figure}

\subsection{Evaluation of Attributes Learning}
The experiments and analysis of the ALN are mainly based on the CompCars dataset and the UTS dataset. The web-nature data in the CompCars dataset are labeled with five viewpoints and twelve types about 136000 and 97500 images, respectively. We neglect those images without type annotation and randomly split the remaining data into the training and validation subsets as 7:3. In addition to pose estimation and type classification, we complement the annotation of five common vehicle colors on about 10000 images for evaluation of color recognition. Besides, for type classification, we compare the results of both the selected 6 common vehicle types and the total 12 types as mentioned in Section 3. 3. Vehicle verification (i.e., binary classification of vehicle and background) is evaluated in all the experiments as well.

In the following subsections, we implement four different experiments to investigate the gain of the multi-task architecture, the accuracy by inputs with different image qualities, the effect of layer depths and the difficulty of fine-grained classification under different viewpoints.

\begin{table}
\center
\newcommand{\tabincell}[2]{\begin{tabular}{@{}#1@{}}#2\end{tabular}}
\caption{\small
\textbf{Evaluation (\%) of attributes annotation for vehicles on the UTS dataset}}
\label{table:t1}
\resizebox{0.975\textwidth}{!}{
\begin{tabular}{|c|c|c|c|c|c|}
\cline{1-6}
\multirow{2}{*}{\textbf{\tabincell{c}{Tasks}}}&\multirow{2}{*}{\textbf{Vehicle Verification}} &\multirow{2}{*}{\textbf{Pose Estimation}}&\multicolumn{2}{c|}{\textbf{\tabincell{c}{Vehicle Type\\ Classification}}}&\multirow{2}{*}{\textbf{Color Recognition}} \\
\cline{4-5}
& & & \tabincell{c}{\textbf{12 types}} & \tabincell{c}{\textbf{6 types}} &\\
\hline
\hline
\multicolumn{6}{|c|}{\textbf{Comparison of single-task learning (STL) and multi-task learning (MTL) for attributes prediction}}\\
\hline
STL &98.73 &96.94 &60.37 &88.32 &78.33 \\
MTL  &\textbf{99.45} &98.03 &69.64 &94.91 &\textbf{79.25 }\\
STL feature+SVM& 99.11&97.12 & 60.86&90.75 & 78.06\\
MTL feature+SVM&99.36 &\textbf{98.10} &\textbf{69.86} &\textbf{95.12} &79.19 \\
\hline
\hline
\multicolumn{6}{|c|}{\textbf{Comparison of Attributes prediction with different sizes of vehicle images}}\\
\hline
$28\times28$&90.45 &83.49 &37.52 &53.66 &49.73 \\
$56\times56$&98.12 &91.33 &52.02 &77.02 &66.14 \\
$112\times112$&99.37 &96.56 &63.41 &90.67 &\textbf{80.31} \\
$224\times224$&\textbf{99.45 }&\textbf{98.03} & \textbf{69.64}&\textbf{94.91} &79.25 \\
\hline
\hline
\multicolumn{6}{|c|}{\textbf{Comparison of Attributes prediction with different deep models}}\\
\hline
ALN based on FVPN ($depth=4$)&95.96 &81.21 &27.26 &43.12 &65.12 \\
ALN based on AlexNet ($depth=8$)&\textbf{99.51} &95.76 &66.01 &89.25 & 77.90\\
ALN based on GoogLeNet ($depth=22$)&99.45 &\textbf{98.03} &\textbf{69.04} &\textbf{94.91} &\textbf{79.25} \\
\hline
\end{tabular}
}
\end{table}

\subsubsection{Single-task learning or multi-task learning? }
We first compare the multi-task ALN with the case of training networks for each attribute separately (i.e., single task).  In addition, results by the combination of deep learned features and an SVM classifier are compared as well. All the model architectures are based on the GoogLeNet, and 1024-dimensional features are extracted from layer \emph{pool5/7$\times7$\_s1} to train the corresponding SVM classifier \cite{CC01a}. As shown in the top part of Table 2, the multi-task model consistently achieves higher accuracies on four different tasks, which reveals the benefit of joint training. Although the combination of extracted features and SVM classifiers sometimes can lead to a small increase, we still prefer the proposed end-to-end model because of its elegance and efficiency.

\subsubsection{How small a vehicle size can DAVE annotate?}
Since vehicles within surveillance video frames are usually in different sizes. Visual details of those vehicles far from the camera are significantly unclear. Although they can be selected by the FVPN with coarse requirements, after rescaling to $224\times224$, these vehicle proposals with low image clarity are hard to be annotated with correct attributes by the ALN. To explore this problem, we test vehicle images with original sizes of 224, 112, 56 and 28 using the trained ALN. The middle part of Table 2 illustrates that the higher resolution the original input size is, the better accuracy it can achieve.

\subsubsection{Deep or shallow?}
How deep of the network is necessary for vehicle attributes learning is also worth to be explored. Since our ALN can be established on different deep models, we compare popular deep networks: AlexNet \cite{krizhevsky2012imagenet} and GoogLeNet with 8 layers and 22 layers, respectively. As VGGNet (16 layers version) \cite{simonyan2014very} configured with numerous parameters requires heavy computation and large memory, we do not expect to employ it for our ALN. Besides, our proposed shallow FVPN with 4 layers is also used for attributes learning. From the bottom part of Table 2, we can see that a deeper network does not obtain much better performance on vehicle verification compared to a shallow one. However, for pose estimation, type classification and color recognition, the deepest GoogLeNet consistently outperforms other nets with obvious gaps. Particularly for type classification which belongs to fine-grained categorization, the shallow FVPN gives extremely poor results. It illustrates that a deeper network with powerful discriminative capability is more suitable for fine-grained vehicle classification tasks.

\subsubsection{Fine-grained categorization in different views.}
Finally, since vehicle type classification belongs to fine-grained categorization, we are interested in investigating its difficulty in different views due to its importance for our future work such as vehicle re-identification. As demonstrated in Table 3, for both 12-type and 6-type classification, higher precision is easier to be achieved from side and rearside views, while it is difficult to discriminate vehicle types from the front view. In other words, if we aim to re-identify a target vehicle from two different viewpoints, the type annotation predicted from a side view is more credible than that from a front view.

\begin{table}
\center
\newcommand{\tabincell}[2]{\begin{tabular}{@{}#1@{}}#2\end{tabular}}
\caption{\small
\textbf{Evaluation (\%) of fine-grained vehicle type classification on the UTS dataset}}
\label{table:t1}
\resizebox{0.65\textwidth}{!}{
\begin{tabular}{|c|c|c|c|c|c|}
\hline
Number of vehicle type&Front &Rear &Side &FrontSide &RearSide  \\
\hline
12&58.02 &60.37 &66.73 &61.28 &64.90  \\
\hline
6&79.42 &84.60 &92.93 &86.77 &92.53  \\
\hline
\end{tabular}
}
\end{table}

Fig. 7 shows the qualitative results of our DAVE evaluated on the UTS vehicle dataset. It demonstrates that our model is robust to detect vehicles and annotate their poses, colors and types simultaneously for urban traffic surveillance. The failure cases mainly take place on incorrect colors and types.

\begin{figure}
  \centering
  \includegraphics[width=0.95\textwidth]{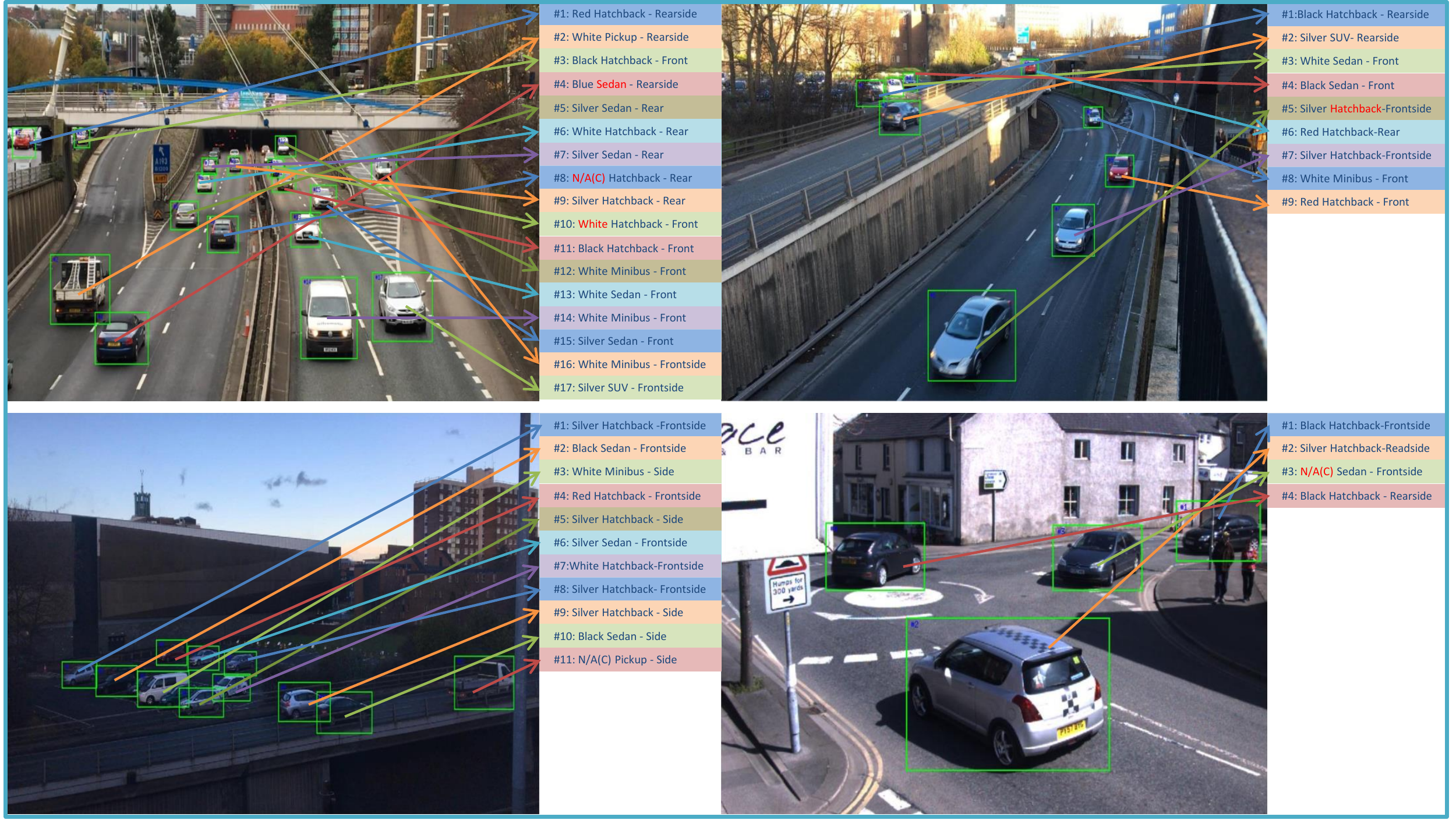}\\
  \caption{\footnotesize
  \textbf{}Qualitative results of attributes annotation. Red marks denote incorrect annotation, and N/A(C) means a catch-all color.}
  \label{exampless}
\end{figure}

\section{Conclusion}
In this paper, we developed a unified framework for fast vehicle detection and annotation: DAVE, which consists of two convolutional neural networks FVPN and ALN. The proposal and attributes learning networks predict bounding-boxes for vehicles and infer their attributes: pose, color and type, simultaneously. Extensive experiments show that our method outperforms state-of-the-art frameworks on vehicle detection and is also effective for vehicle attributes annotation. In our on-going work, we are integrating more vehicle attributes such as make and model into the ALN with high accuracy, and exploiting these attributes to investigate vehicle re-identification tasks.

\bibliographystyle{splncs}
\bibliography{eccv}
\end{document}